\newcommand{\etal}{\textit{et al. }}
\documentclass[lettersize,journal]{IEEEtran}
\usepackage{amsmath,amsfonts}
\usepackage{amssymb}
\usepackage{algorithmic}
\usepackage{algorithm}
\usepackage{array}
\usepackage[caption=false,font=normalsize,labelfont=sf,textfont=sf]{subfig}
\usepackage{textcomp}
\usepackage{stfloats}
\usepackage{url}
\usepackage{verbatim}
\usepackage{graphicx}
\usepackage{cite}
\usepackage{threeparttable}
\usepackage{booktabs}

\usepackage{cases}
\usepackage{multirow}
\usepackage[colorlinks,linkcolor=black,citecolor=black,urlcolor=black]{hyperref}

\begin{document}
%
\title{PDMP: Rethinking Balanced Multimodal Learning via Performance-Dominant Modality Prioritization}
%
%
%

\author{Shicai Wei, Chunbo Luo,~\IEEEmembership{Senior Member,~IEEE,}, Qiang Zhu,
        Yang Luo

\thanks{Shicai Wei is with the Laboratory of Intelligent Collaborative Computing, University of Electronic Science and Technology of China. 

Yang Luo, and Chunbo Luo are with the School of Information and Communication Engineering, University of Electronic Science and Technology of China, Sichuan, China. (\textit{Corresponding author: Chunbo Luo.})

Qiang Zhu is with Peng Cheng Laboratory, Shenzhen, China.}
        
        }


%
%

\markboth{Journal of \LaTeX\ Class Files,~Vol.~14, No.~8, August~2015}%
{Shell \MakeLowercase{\textit{et al.}}: Bare Demo of IEEEtran.cls for IEEE Journals}
%

\maketitle


\begin{abstract}
Multimodal learning has attracted increasing attention due to its practicality. However, it often suffers from insufficient optimization, where the multimodal model underperforms even compared to its unimodal counterparts. Existing methods attribute this problem to the imbalanced learning between modalities and solve it by gradient modulation. This paper argues that balanced learning is not the optimal setting for multimodal learning. On the contrary, imbalanced learning driven by the performance-dominant modality that has superior unimodal performance can contribute to better multimodal performance. And the under-optimization problem is caused by insufficient learning of the performance-dominant modality. To this end, we propose the Performance-Dominant Modality Prioritization (PDMP) strategy to assist multimodal learning. Specifically, PDMP firstly mines the performance-dominant modality via the performance ranking of the independently trained unimodal model.  Then PDMP introduces asymmetric coefficients to modulate the gradients of each modality, enabling the performance-dominant modality to dominate the optimization. Since PDMP only relies on the unimodal performance ranking, it is independent of the structures and fusion methods of the multimodal model and has great potential for practical scenarios. Finally, extensive experiments on various datasets validate the superiority of PDMP. 
\end{abstract}


\section{Introduction}

Multimodal learning has attracted growing interest in many fields, such as action recognition~\cite{ar1,Wei2023OneStageModalityDistillation,Wei2023MMANet,Wei2024ScaledDecoupledDistillation}, object detection~\cite{mm_detection2,mm_detection3}, and audiovisual emotion recognition~\cite{tmm-c1,tmm-c2,tmm-c3}. While it holds significant potential for improving model robustness and performance, the heterogeneity of multimodal data complicates the effective use of cross-modal correlations and complementarities. Recent studies~\cite{ogm,umt,pmr} indicate that although multimodal approaches generally surpass unimodal ones, their performance remains suboptimal. In many cases, some modalities are not fully learned or used during training, which limits the final results. 


Existing research attributes this limitation to modality imbalance, where the optimization-dominant modality that converges faster suppresses weaker ones, hindering the full utilization of multimodal information~\cite{wh,ogm,umt,pmr,agm,MLA}. To address this, some methods strengthen the weak modalities using auxiliary components, such as unimodal classifiers~\cite{wh}, pre-trained models~\cite{umt}, or resampling techniques~\cite{FMV}. However, these strategies often introduce additional complexity. As a simpler alternative, several works~\cite{ogm,pmr,agm} modulate gradients to balance learning across modalities by identifying the optimization-dominant modality and adjusting its gradient accordingly. A recent method further transforms joint multimodal learning into an alternating unimodal training process to reduce inter-modality interference~\cite{MLA}.


\begin{figure*}[ht]
\centering
\includegraphics[width=1.0\textwidth]{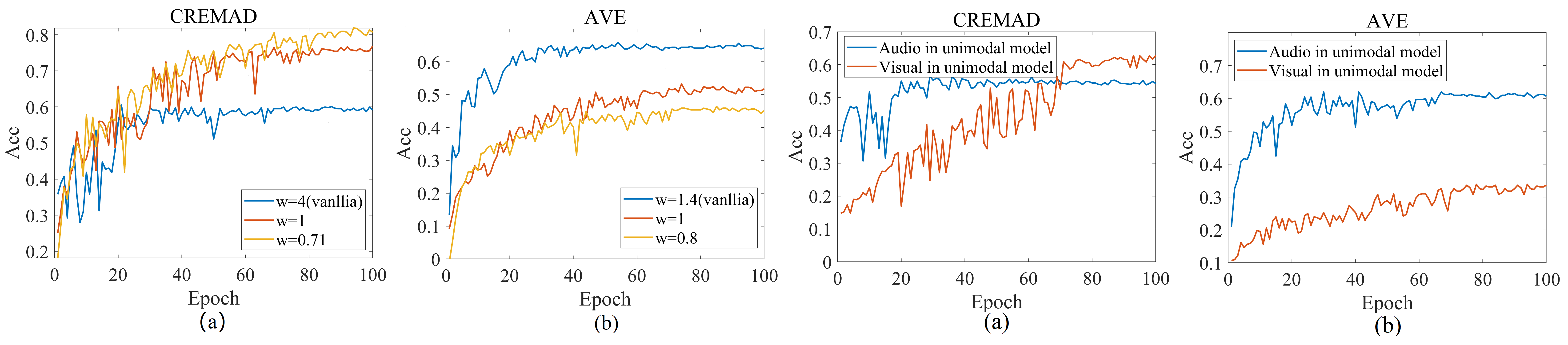} 
\caption{ The visualization on two audio-visual datasets: CREMA-D and AVE. (a) and (b) present the performance curves of the multimodal models under different values of $ w$. $w$ is the optimization dependency coefficient (see Eq.\ref{impact_new}), which measures the relative contribution of the audio and visual modalities during training. `vanilla' means the default $w$ without any gradient modulation. (c) and (d) present the unimodal performance curves.}
\label{diveristy-compare}
\end{figure*}



While these methods have shown promising results, they all share a common assumption that sub-optimal multimodal performance is caused solely by imbalanced learning between modalities. In this paper, we challenge this assumption. As shown in Figure~\ref{diveristy-compare}, balancing optimization between audio and visual modalities ($w=1$) improves multimodal performance on the CREMA-D dataset, but surprisingly decreases it on the AVE dataset. More notably, even on CREMA-D, the best performance occurs not at the balanced point ($w=1$), but at an imbalanced setting ($w=0.71$). These observations suggest that balanced learning is not always optimal, driving us to further explore the real bottleneck that limits multimodal learning performance.



Accordingly, we reframe the conventional joint multimodal learning process by transforming it into the ensembling learning of multiple unimodal processes. From an information-theoretic perspective, we prove that the multimodal model should pay more attention to the performance-dominant modality. This helps achieve the best information capacity. For example, in CREMA-D, the visual modality dominates in unimodal performance. Thus, existing balanced learning methods that accelerate the learning of the visual modality and suppress the learning of the audio modality can lead to better multimodal results. However,  on the AVE dataset, the audio modality is dominant. Thus, balanced learning is against the criteria of optimal information capacity, thereby declining the multimodal performance on the AVE dataset.


To this end, we propose the Performance-Dominant Modality Prioritization (PDMP) strategy to assist multimodal learning. Specifically, PDMP mines the performance-dominant modality via the performance ranking of the independently trained unimodal model. Then, PDMP introduces asymmetric coefficients to modulate the gradients of each modality, encouraging the performance-dominant modality to dominate the optimization when accelerating the optimization of each modality. In detail, for the performance-dominant but optimization-weak modality (e.g., vision in CREMA-D), PDMP will boost its gradient and suppress others to enhance reliance on it. For the performance-dominant and well-optimized modality (e.g., audio in AVE), PDMP equally scales all gradients to maintain its dominance while speeding up optimization.  Finally, we evaluate PDMP in various multimodal tasks on different datasets, confirming its effectiveness and versatility.

Overall, our contributions are summarized as follows,


\begin{itemize}
    \item We challenge the common assumption that balanced learning leads to optimal multimodal performance, and identify insufficient learning of the performance-dominant modality as the key bottleneck. This provides an insightful view to study under-optimized multimodal learning.
    
    \item  We prove that imbalanced optimization dominated by the performance-dominant modality can contribute to better performance, offering theoretical insights into effective multimodal optimization.

    
    \item  We propose PDMP, a plug-and-play strategy that prioritizes the performance-dominant modality in multimodal learning via unimodal ranking and asymmetric gradient modulation.

    \item Extensive experiments on various tasks and datasets demonstrate the effectiveness and generalization of the proposed PDMP in boosting multimodal learning.


\end{itemize}

\section{Related Works}
\subsection{Multimodal Learning}
Multimodal learning frameworks that integrate information from diverse data sources have demonstrated significant performance advantages compared to single-modality approaches across various domains~\cite{ar1,Wei2024GradientDecoupledLearning,Wei2024PrivilegedModalityLearning,Wei2023DiversityGuidedDistillation,Wei2023MSHNet}. Significant research efforts have been directed toward leveraging multimodal information to achieve state-of-the-art performance in specialized applications. These include advanced action recognition systems incorporating motion and temporal features~\cite{MH3,MARS,MH2}, RGB-D based semantic segmentation architectures~\cite{rgbd_seg1,rgbd_seg2,rgbd_seg3}, and integrated audio-visual speech recognition models~\cite{av1,av2}. However, a critical challenge persists in current multimodal approaches: conventional joint training paradigms often fail to adequately optimize individual modality representations while learning cross-modal interactions. This limitation results in suboptimal model performance, where in some cases, multimodal systems may not surpass the accuracy of well-trained single-modality deep neural networks~\cite{wh}. Recent studies suggest that this performance gap may stem from inadequate attention to unimodal feature learning during the joint optimization process.

\subsection{Under-optimization Multimodal Learning}

The limitations observed in previous multimodal learning approaches have prompted researchers to investigate the underlying causes of these issues. One recent study~\cite{wh} identified that different modalities tend to overfit and generalize at varying speeds, which leads to suboptimal performance when trained together using a unified optimization scheme. To address this, the authors compute the logit outputs of individual modalities along with their fused representations and refine the gradient mixing mechanism to determine more effective weighting strategies for each modality branch.

In addition, a number of works~\cite{umt,ogm,pmr,mmcosine,agm,MLA,FMV} have revealed that when modalities are trained jointly, the stronger-performing modality often dominates the gradient updates, which suppresses the learning progress of the weaker modality. To tackle this issue, the UMT framework~\cite{umt} focuses on improving unimodal learning by leveraging knowledge distillation from pretrained networks. However, this approach introduces greater architectural complexity and computational overhead. Therefore, the gradient modulation methods enlarge the gradient of the weaker modality in multimodal learning to balance the optimization of different modality encoders directly.  Specifically, OGM~\cite{ogm} proposes on-the-fly gradient modulation to manage the optimization of each modality adaptively. MMCosine~\cite{mmcosine} performs modality-wise L2 normalization to features and weights towards balanced and better multi-modal fine-grained learning. AGM~\cite{agm} introduces an adaptive gradient modulation method that can boost the performance of multimodal models with various fusion strategies. CGGM~\cite{guo2024classifier} balances multimodal learning by both considering the magnitude and
direction of the gradients. LFW~\cite{LFM} dynamically integrates unsupervised contrastive learning and supervised multimodal
learning to address the modality imbalance problem. GMML~\cite{zhang2025gmml} suppresses dominant modality gradients while proportionally amplifying weaker ones simultaneously. GGDM~\cite{hu2025geometric} leverages the volumes of gradient polyhedra to perform gradient modulation, encouraging alignment among gradients and promoting a synergistic optimization effect. While gradient modulation methods show good results, improving weak modalities often degrades the performance of strong ones due to the inter-modality conflict. However, these methods directly reduce the gradient magnitude of the dominant modality, aiming to lessen its suppressive influence on others.

Building upon this,  the collaborative optimization methods~\cite{MLA,mmpareto,reconboost, diagnosing,dgl} are proposed to improve the unimodal learning of each modality, including the domainaint one, in multimodal learning. MLA~\cite{MLA} transforms the conventional joint multimodal learning process into an alternating unimodal learning process to minimize inter-modality interference directly. ReconBoost~\cite{reconboost} updates a fixed modality each time via a dynamical learning objective to overcome the competition with the historical models. MMPareto~\cite{mmpareto} leverages the Pareto integration technique to catch innocent unimodal assistance, avoiding its conflict with multimodal optimization. Besides, Wei~\etal~\cite{diagnosing} propose the Diagnosing \& Re-learning method to overcome the intrinsic limitation of modality capacity via the network re-initialization technique. DGL~\cite{dgl} decouples the optimization of the modality encoder and modality fusion module in the multimodal model, eliminating optimization conflicts among different modality encoders. ARL~\cite{arl} proposes to pay more attention to the high-performance modality.

\begin{figure*}[ht]
\centering
\includegraphics[width=0.80\textwidth]{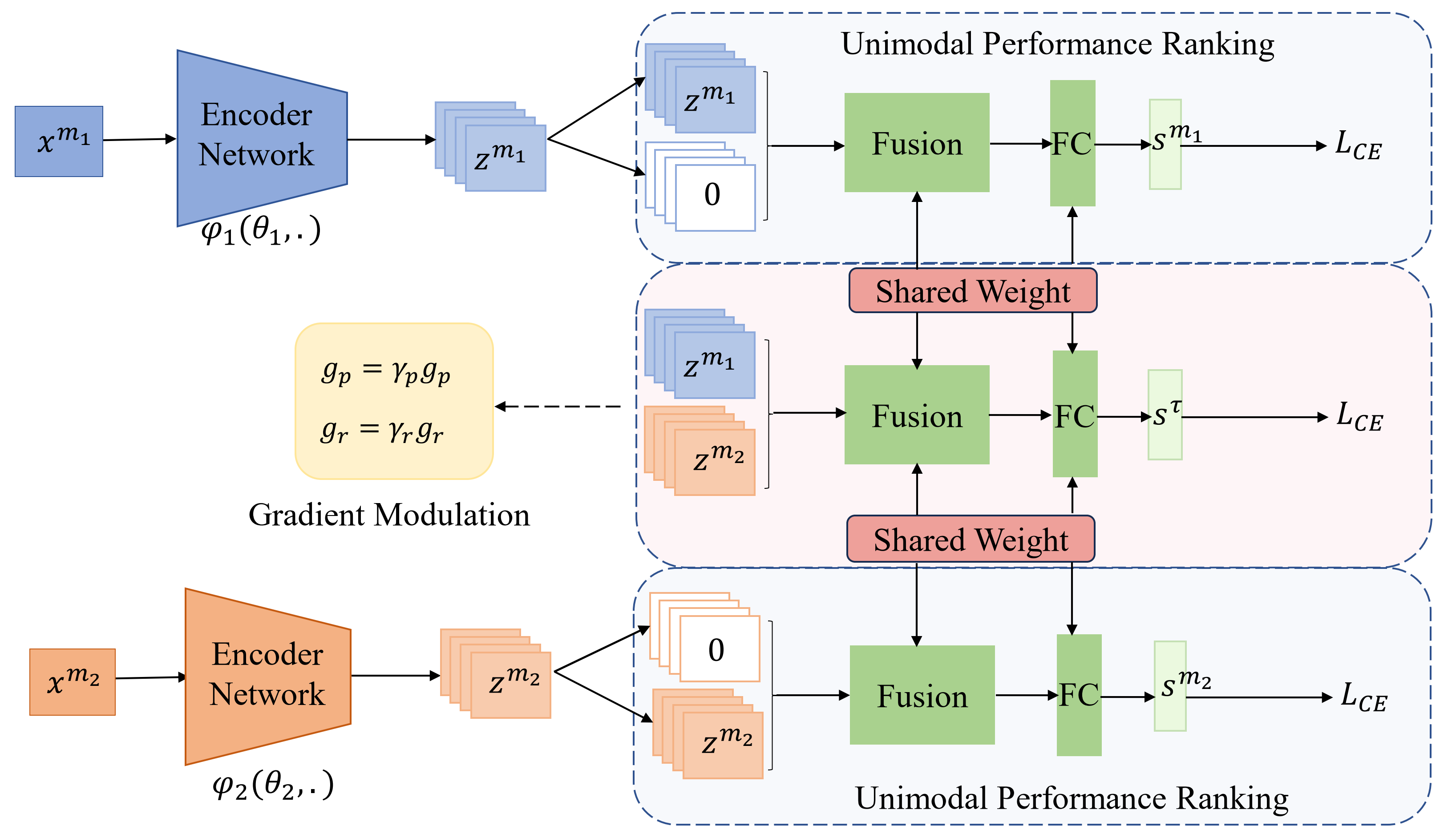}
\caption{ Illustration of back-propagation processing for balanced learning and the proposed performance-dominant modality prioritization. Balanced learning mines the optimization-dominant modality and aligns its gradient $g_o$ with the gradient  $g_r$ of the remaining modality.  In contrast, performance-dominant modality prioritization mines the performance-dominant modality and makes its gradient $g_p$ larger than the gradient  $g_r$ of the remaining modality, guiding it to dominate the optimization.}
\label{framework}
\end{figure*}

\section{Methods}

We first reanalyze the under-optimization phenomenon in multimodal learning and provide an analytical perspective on how modality-dependent optimization strength should be allocated according to their relative predictive reliability. Based on this analysis, we further motivate the PDMP strategy, which prioritizes the performance-dominant modality during optimization.

\subsection{Under-optimization Analysis in Multimodal Learning}
\label{anas}

\textbf{Multimodal learning model}. 
We consider a dual-modality classification setting with inputs from modality $m_1$ and $m_2$. 
The dataset is denoted as $\mathcal{D}=\{(x_i^{m_1}, x_i^{m_2}, y_i)\}_{i=1}^N$, where $y_i \in \{1,\ldots,M\}$ indicates the class label.

Each modality is encoded by a modality-specific encoder $\phi_1(\theta_1,\cdot)$ and $\phi_2(\theta_2,\cdot)$, producing representations
\begin{equation}
\boldsymbol{z}_1 = \phi_1(\theta_1, x_i^{m_1}), \quad
\boldsymbol{z}_2 = \phi_2(\theta_2, x_i^{m_2}).
\end{equation}

The representations are fused by a fusion function $\phi_f(\theta_f,\cdot)$ and fed into a linear classifier:
\begin{equation}
\boldsymbol{z}_f = \phi_f(\theta_f, \boldsymbol{z}_1, \boldsymbol{z}_2), \quad
f(x_i) = \boldsymbol{W}\boldsymbol{z}_f + \boldsymbol{b}.
\end{equation}

Here, take the most widely used vanilla fusion method, concatenation, as an example, $\boldsymbol{z}_f=[\boldsymbol{z}_0;\boldsymbol{z}_1]$ and thus $f(x_i)$ can be rewritten as follows,

\begin{equation}
\label{decouple_new}
f(x_i) = \boldsymbol{W}_1 \boldsymbol{z}_1 + \boldsymbol{b}_1 + \boldsymbol{W}_2 \boldsymbol{z}_2 + \boldsymbol{b}_2
\end{equation}
where $\boldsymbol{W} = [\boldsymbol{W}_1; \boldsymbol{W}_2]$, with $\boldsymbol{W}_1 \in \mathbb{R}^{M \times d_1}$, $\boldsymbol{W}_2 \in \mathbb{R}^{M \times d_2}$, and $\boldsymbol{b}_1, \boldsymbol{b}_2 \in \mathbb{R}^M$. We then define the modality-specific logit outputs as follows,
\begin{equation}
\label{s}
{s_1} = \boldsymbol{W}_1 \boldsymbol{z}_1 + \boldsymbol{b}_1, \quad {s_2} = \boldsymbol{W}_2 \boldsymbol{z}_2 + \boldsymbol{b}_2
\end{equation}


such that the final prediction is the additive combination $\boldsymbol{s}_1 + \boldsymbol{s}_2$.

\textbf{Optimization dependency}. 
Since the back-propagated gradients of each modality are directly influenced by the magnitude of its logit contribution~\cite{ogm}, we can introduce a simple logit proxy to characterize the relative optimization dependency between modalities,
\begin{equation}
\label{impact_new}
w = \frac{w_1}{w_2}= \frac{\sum_{j=1}^M |s_1(j)|}{\sum_{j=1}^M |s_2(j)|},
\end{equation}
where $| \cdot |$ denotes the absolute value.
A larger $w$ indicates stronger optimization influence from modality $m_1$, while $w=1$ corresponds to equal logit-level contribution.
Previous studies~\cite{ogm,pmr,agm} implicitly or explicitly enforce $w=1$, assuming balanced optimization leads to optimal multimodal learning.
However, empirical evidence (Fig.~\ref{diveristy-compare}) suggests that such enforced balance is not always optimal, motivating a deeper analysis of the desired optimization dependency.

\textbf{Analytical motivation for unequal optimization dependency}. 
Rather than enforcing equal contribution, we aim to understand how modality contributions should be weighted according to their predictive relevance.
From an information-theoretic perspective, a desirable fusion strategy should maximize the mutual information between model outputs and ground-truth labels.

To explicitly control the contribution of each modality, we introduce non-negative scaling factors $u_1$ and $u_2$ and consider the following objective:
\begin{equation}
\label{eq:mi_obj}
\max_{u_1,u_2} I(u_1 s_1 + u_2 s_2;\, y).
\end{equation}

This formulation does not impose any assumption on the distributions of $s_1$ and $s_2$, nor does it require them to be independent.
Instead, it provides a general information-theoretic perspective on how modality contributions affect predictive performance.

Recall that the mutual information between the model output $z=u_1 s_1 + u_2 s_2$ and the label $y$ can be written as
\begin{equation}
I(z;y) = H(y) - H(y|z),
\end{equation}
where $H(y)$ is the entropy of the labels and does not depend on $u_1$ or $u_2$.
Therefore, maximizing $I(z;y)$ is equivalent to minimizing the conditional entropy $H(y|z)$.

The conditional entropy admits the following variational form:
\begin{equation}
H(y|z) = \inf_{q(y|z)} 
\mathbb{E}_{p(y,z)} \left[ -\log q(y|z) \right],
\end{equation}
where the infimum is achieved when $q(y|z) = p(y|z)$.
Therefore, minimizing $H(y|z)$ corresponds to finding a representation $z$ that allows the most accurate prediction of $y$. Let $\mathcal{M}_i:= I(s_i;y)$ denote a generic discriminative measure of modality $m_i$. By the data processing inequality and the monotonicity of conditional entropy with respect to informative projections~\cite{cover_thomas,tishby1999ib}, we can get the following result,

\begin{equation}
\label{eq:u_monotonic}
\frac{u_1}{u_2}
\;\propto\;
\!\ \frac{\mathcal{M}_1}{\mathcal{M}_2} ,
\quad
\end{equation}

Consequently, the optimal modality contribution should allocate larger weights to modalities with stronger predictive relevance.
This analysis suggests that the optimal ratio $u_1/u_2$ should be monotonically increasing with the relative discriminative strength of the two modalities:

\textbf{Explanation of the performance gain and decline of existing balanced learning methods on the CREMA-D and AVE datasets}. As discussed, the multimodal model should pay more attention to the modality with superior mutual information with the label. In the CREMA-D dataset, the visual modality performs better than the audio modality. Therefore, according to the optimal mutual information criterion, the multimodal mode should pay more attention to the visual modality when making the decision. However, traditional multimodal models often focus too much on audio and suppress the visual modality. Therefore, we should slow down the optimization rate of the audio modality and speed up the optimization rate of the visual modality. This operation is consistent with existing methods that aim to balance the learning of audio and visual modalities. Therefore, existing balanced learning methods can achieve the performance gain in the CREMA-D dataset. Nevertheless, they fail to allow the multimodal model to refer more to the visual modality. Consequently, we can observe additional performance gains by further amplifying the optimization rate of the visual modality and making the model’s reliance on visual cues over audio.

In contrast, the audio modality is performance-dominant in the AVE dataset, and traditional multimodal models naturally exhibit a higher optimization bias for the audio modality~\cite{ogm,mmpareto}. Therefore, we maintain or even strengthen the model’s decision dependency on the audio modality to meet the optimal mutual information criterion. However, existing balanced learning methods also suppress the optimization of audio and promote the optimization of the visual modality to equalize their contributions. This approach contradicts the optimal mutual information criterion, thereby leading to suboptimal performance.

\subsection{Performance-Dominant Modality Prioritization}

As discussed, balancing the learning of different modalities is not the optimal setting for multimodal learning. On the contrary, we should guide the performance-dominant modality to dominate the optimization of multimodal learning. To this end, we propose a simple but effective performance-dominant modality prioritization (PDMP) strategy to improve the performance of multimodal learning. As shown in Algorithm~\ref{dgd}, it consists of two stages: modality analysis and gradient modulation.

\begin{figure}[t]
\centering
\includegraphics[width=0.5\textwidth]{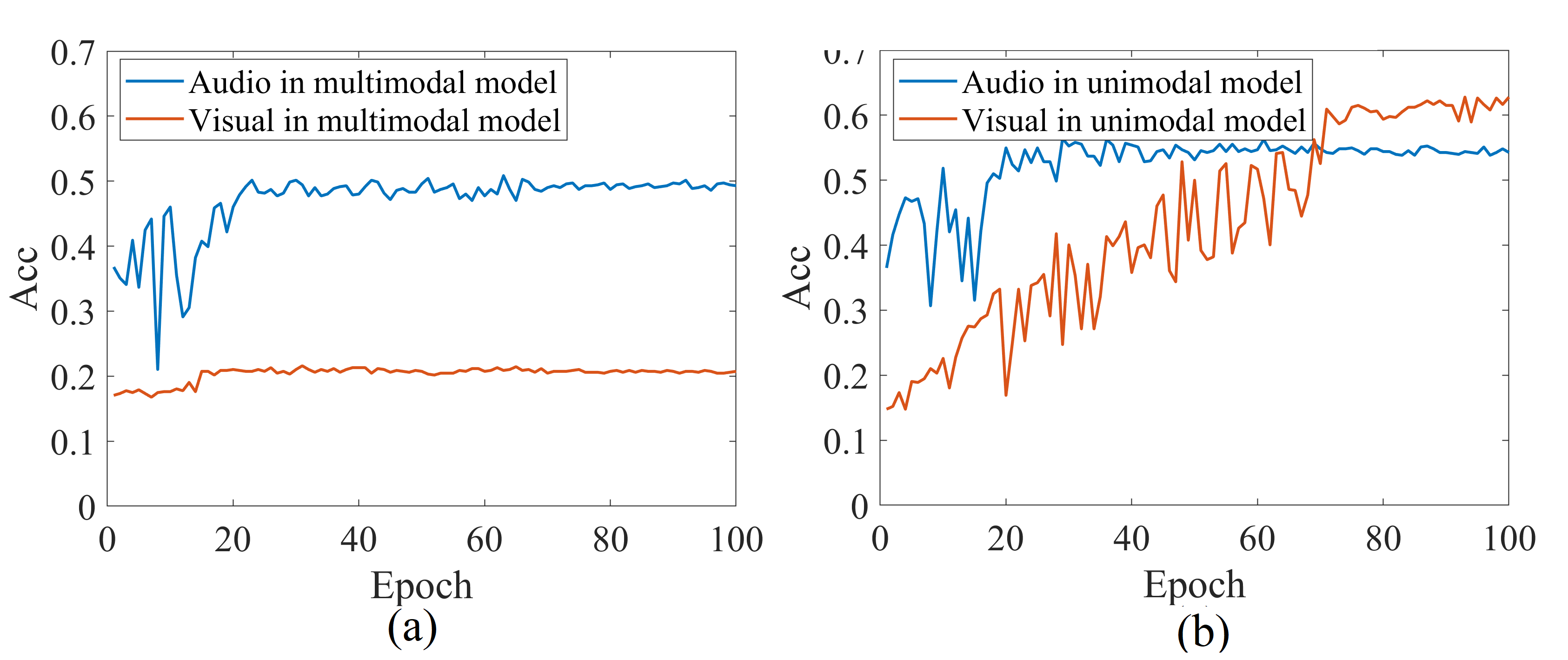} 
\caption{ The visualization on the CREMA-D datasets. (a) presents the performance of the audio and visual branches in the multimodal model. (b) presents the performance of the audio and visual modality in the unimodal model.  }
\label{dd}
\end{figure}

\textbf{Modality Analysis} This stage determines the performance-dominant modality that has superior unimodal performance and the optimization-dominant modality that has superior branch performance in the multimodal model. As shown in Fig.~\ref{dd} (a), the audio modality outperforms the visual modality in the multimodal model on the CREMA-D dataset. However, as shown in Fig.~\ref{dd} (b), the visual modality achieves better performance than the audio modality in the unimodal model on the CREMA-D dataset. This shows that the performance-dominant modality is not equal to the optimization-dominant modality, and we can not determine them via the conventional branch performance ranking in the multimodal model.


To this end, we propose to analyze the modality property via unimodal performance ranking. Specifically, we train the unimodal model for each modality and rank their performance. According to the definition, the modality with better performance at the end of training is the performance-dominant modality. Besides, according to the memorization effect~\cite{mem1,mem2,mem3} that DNNs tend to first memorize simple examples before hard examples, the modality with better performance at the beginning of training will be the optimization-dominant modality. This is because the modality with better performance at the beginning will contribute to a higher gradient in the early training and then dominate the optimization of the multimodal model.

In addition, to reduce the computational burden of training extra unimodal models, we propose to analyze the modality property in a subset. The details can be seen in Section~\ref{ab}.

\textbf{Gradient Modulation} As shown in Fig.~\ref{framework}, let $m_{p}$ and  $m_{r}$ denote the determined performance-dominant modality and the remaining modality, respectively. We introduce two hyper-parameters $\gamma_{p}$ and $\gamma_{r}$ to accelerate their optimization and enable the performance-dominant modality to dominate the optimization. Specifically, the modulated gradient $g_{p}$ for modality $m_{p}$ will be updated as follows,

\begin{equation}
\label{update_mp}
g_{p}=\gamma_{p} {g_p}^{'}
\end{equation} where ${g_p}^{'}$ denotes the original gradient modality $m_{p}$.



And  the gradient $g_{r}$ for the modality $m_r$ will be updated as follows,

\begin{equation}
\label{update_mr}
g_{r}=\gamma_{r} {g_r}^{'}
\end{equation} where ${g_r}^{'}$ denotes the original gradient modality $m_{r}$.


According to the optimal mutual information criterion, it is necessary to improve the gradient of the performance-dominant modality to accelerate its learning. Therefore, $\gamma_{p}$ must be larger than $1$. Then, if the remaining modality $m_{r}$ is the optimization-dominant modality, we need to slow down its optimization so that $m_{p}$ can dominate the optimization in multimodal learning. Thus, $m_{r}$ must be smaller than $1$, for simplicity, we set it as $\frac{1}{\gamma_{p}}$. Besides, if $m_{r}$ is not the optimization-dominant modality, like the visual modality in the AVE dataset, we can also improve the gradient of $m_{r}$ to accelerate its optimization. This will not affect the dominance of $m_{p}$ in the multimodal model. Therefore, $\gamma_{r}$ can be determined as follows for simplicity, 

\begin{equation}
\label{gammar}
\gamma_{r}=\left\{\begin{array}{cl}
\frac{1}{\gamma_{p}}  & m_{p} \neq m_{o} \\
\gamma_{p} & m_{p} = m_{o}
\end{array}\right.
\end{equation} where $m_{o}$ denotes the optimization-dominant modality.

\begin{algorithm}[t]
	\caption{Multimodal learning with PDMP strategy} 
	\label{dgd} 
	\begin{algorithmic}
		\REQUIRE Training dataset D, iteration number T, hyperparameter $\gamma_p$.
        \STATE  \textbf{Modality Analysis}:
        \STATE Train the unimodal model for each input modality
within T iterations;
        \STATE  Determine the performance-dominant modality
and the remaining modality via unimodal performance ranking;
         \STATE Determine optimization-dominant modality  via modality branch
performance ranking in the multimodal;
        \STATE Determine the parameter $\gamma_r$ via Eq.\ref{gammar};

        \STATE  \textbf{Gradient Modulation}:
        
        \STATE \textbf{for} t = 0, ... , T - 1 \textbf{do}
        \STATE \quad\quad Feed-forward the batched data to the model.

        \STATE \quad\quad Calculate backward gradient.
        \STATE \quad\quad Modify the gradient of each modality via Eq.\ref{update_mp} and \ref{update_mr}; 
        \STATE \quad\quad Update model parameters via the modified gradient.
	\end{algorithmic} 
\end{algorithm}

Here, we discuss PDMP with existing balanced multimodal learning methods again to clarify its contribution. Specifically, existing balanced methods aim to balance the training process of different modalities, but PDMP guides the performance-dominant modality to dominate the training process. The paper proves theoretically and experimentally that the latter can achieve better results (see Section~\ref{anas}). In detail, the balanced methods increase the gradient of visual modality in the CREMA-D dataset to ensure the optimization dependency coefficient $w$ is equal to 1. However, PDMP aims to ensure the optimization dependency coefficient $w$ is equal to the performance ratio demonstrated in Eq.12. As shown in Figure~\ref {diveristy-compare}(a) and Table~\ref{compare}, PDMP can achieve better performance than existing balanced methods.

\section{Experiments}
\subsection{Datasets}
\textbf{CREMA-D}~\cite{Crema} is a multimodal dataset designed for emotion recognition, incorporating both audio and visual information. It contains 7,442 video clips labeled with six common emotion categories. We randomly partition the dataset into 6,698 samples for training and 744 for testing.

\textbf{AVE}~\cite{AVE} is an audio-visual video dataset for audio-
visual event localization. It contains 4,143 10-second videos for 28 event classes. Here, we extract the frames from event-localized video segments and capture the audio clips within the same segments, constructing a labeled multimodal classification dataset. The training and validation split of the dataset follows~\cite{AVE}.

\textbf{Kinetics-Sounds (KS)}~\cite{ks} is a dataset formed by filtering the Kinetics dataset for 34 human action classes. Each class is chosen to be potentially manifested visually and aurally. This dataset contains 19k 10-second video clips (15k training, 1.9k validation, 1.9k test).

\textbf{CEFA}~\cite{cefa} is a multimodal dataset build for face anti-spoofing task. It consists of three modalities: RGB, Depth, and IR. Here, we conduct experiments on the CEFA dataset to demonstrate the generalization ability of PDMP in the scene using more than two modalities. We follow the cross-ethnicity and cross-attack protocol suggested by the authors and divide it into train, validation, and test sets with 35k, 18k, and 54k samples, respectively.

\textbf{UCF-101}~\cite{soomro2012ucf101} is a dataset for action recognition. It includes RGB videos and optical flow data. There are 101 different actions. The standard split has 9,537 videos for training and 3,783 for testing.

\textbf{VGGSound}~\cite{vggsound} is a video dataset with 309 categories, capturing various audio events in daily life. For our experiment, we employed 168,618 videos for training and validation, and 13,954 videos for testing. This helps study the effectiveness of ARL to the large-scale dataset.



\begin{table*}[t]
\centering
\caption{Comparison with existing modulation strategies on CREMA-D, Kinetics-Sounds, and AVE datasets. The proposed PDMP achieves the best performance. Bold and underline mean the best and second-best results, respectively.}
\label{compare}
\renewcommand\arraystretch{1.1}
\setlength{\tabcolsep}{4mm}{
\begin{tabular}{c|cc|cc|cc}
\hline
Dataset       & \multicolumn{2}{c|}{CREMA-D}                         & \multicolumn{2}{c|}{KS}                              & \multicolumn{2}{c}{AVE}                             \\ \hline
Methods       & \multicolumn{1}{c}{Acc}            & macro F1       & \multicolumn{1}{c}{Acc}            & macro F1       & \multicolumn{1}{c}{Acc}            & macro F1       \\ \hline
Audio-only    & \multicolumn{1}{c}{57.27}          & 57.89          & \multicolumn{1}{c}{48.67}          & 48.89          & \multicolumn{1}{c}{62.16}          & 58.54          \\ 
Visual-only   & \multicolumn{1}{c}{62.17}          & 62.78          & \multicolumn{1}{c}{52.36}          & 52.67          & \multicolumn{1}{c}{31.40}          & 29.87          \\ \hline
Concatenation & \multicolumn{1}{c}{58.83}          & 59.43          & \multicolumn{1}{c}{64.97}          & 65.21          & \multicolumn{1}{c}{66.15}          & 62.46          \\
Grad-Blending & \multicolumn{1}{c}{68.81}          & 69.34          & \multicolumn{1}{c}{67.31}          & 67.68          & \multicolumn{1}{c}{67.40}          & 63.87          \\ 
OGM-GE        & \multicolumn{1}{c}{64.34}          & 64.93          & \multicolumn{1}{c}{66.35}          & 66.76          & \multicolumn{1}{c}{65.62}          & 62.97          \\ 
AGM           & \multicolumn{1}{c}{67.21}          & 68.04          & \multicolumn{1}{c}{65.61}          & 65.99          & \multicolumn{1}{c}{64.50}          & 61.49          \\ 
PMR           & \multicolumn{1}{c}{65.12}          & 65.91          & \multicolumn{1}{c}{65.01}          & 65.13          & \multicolumn{1}{c}{63.62}          & 60.36          \\ 
MMPareto      & \multicolumn{1}{c}{70.19}          & 70.82          & \multicolumn{1}{c}{69.13}          & 69.05          & \multicolumn{1}{c}{68.22}          & 64.54          \\ 
MLA           & \multicolumn{1}{c}{73.21}          & 73.77          & \multicolumn{1}{c}{\underline{69.62}}          & \underline{69.98}          & \multicolumn{1}{c}{\underline{70.92}}          & \underline{67.23}          \\
D\&R          & \multicolumn{1}{c}{\underline{73.52}}          & \underline{73.96}          & \multicolumn{1}{c}{69.10}          & 69.36          & \multicolumn{1}{c}{69.62}          & 64.93          \\ \hline
PDMP           & \multicolumn{1}{c}{\textbf{80.21}} & \textbf{80.34} & \multicolumn{1}{c}{\textbf{72.18}} & \textbf{72.67} & \multicolumn{1}{c}{\textbf{71.28}} & \textbf{67.49} \\ \hline
\end{tabular}}

\end{table*}

\subsection{Experimental settings}

\textbf{Implementation details}. To ensure a fair comparison, we adopt ResNet-18 as the encoder backbone for all evaluated methods on the CREMA-D, AVE, Kinetics-Sounds, and CEFA datasets, following prior works~\cite{ogm,pmr}. For CREMA-D, one frame is extracted from each video clip and resized to 224×224 as the visual input, while the corresponding audio is converted into a spectrogram of size 257×299 using Librosa~\cite{librosa}. For AVE and Kinetics-Sounds, three frames are uniformly sampled from each video clip and resized to 224×224 as visual inputs; the entire audio clip is transformed into a spectrogram of size 257×1,004. For CEFA, both RGB, Depth, and IR modalities are resized to 112×112 and used as inputs to their respective encoders.

All models are trained under consistent settings as in previous studies~\cite{ogm,pmr}, using a mini-batch size of 64, stochastic gradient descent (SGD) with a momentum of 0.9, a learning rate of 1e-3, and a weight decay of 1e-4.

\textbf{Comparison settings}. To evaluate the effectiveness of our proposed PDMP, we compare it against three gradient modulation methods (OGM-GE~\cite{ogm}, AGM~\cite{agm}, and PMR~\cite{pmr}) and four unimodal regularization approaches (G-Blending~\cite{wh}, MLA~\cite{MLA}, MMPareto~\cite{mmpareto}, and D\&R~\cite{diagnosing}). For a fair comparison, all methods share the same ResNet-18 backbone and use feature concatenation for modality fusion.

Note that the original MLA adopts 150 training epochs and a batch size of 16; we align these settings with other baselines by using 100 epochs and a batch size of 64. Moreover, the default learning rate for MMPareto~\cite{mmpareto} and D\&R~\cite{diagnosing} is 2e-3; we adjust it to 1e-3 for consistency across all methods.

Here, we use accuracy and macro F1 score as the metric.

\subsection{Comparison on multimodal tasks}


\textbf{Comparison with other modulation strategies}. Table ~\ref{compare} comprehensively compares the proposed PDMP strategy and other modulation approaches across the CREMA-D, Kinetics-Sounds, and AVE datasets. As we can see, PDMP achieves the best results across all datasets, with clear improvements over existing methods. In particular, it surpasses the second-best approach by 6.25\%, 2.56\%, and 2.18\% on CREMA-D, Kinetics-Sounds, and AVE, respectively. This demonstrates its strong effectiveness.
 
Notably, while existing balance-based methods can improve the performance on the CREMA-D and KS datasets, they decrease the performance on the AVE dataset. This is because the balance-based methods could hinder the learning of performance-dominant modality (audio modality) when they balance the optimization of audio and visual modalities.  In contrast, the proposed PDMP also improves the performance on the AVE dataset by 6.03\% compared to the concatenation baseline.  These results demonstrate the effectiveness of guiding the performance-dominant modality to dominate multimodal learning. Although methods such as MLA and D\&R enhance the learning of audio and visual encoders through unimodal loss, they still do not change the essence of balanced learning. Therefore, their improvement is limited compared to PDMP in the AVE dataset.

\begin{table}[]
\centering
\renewcommand\arraystretch{1.1}
\caption{\textbf{(Left)}: Comparison with imbalanced multimodal learning methods on CEFA dataset with three modalities.\textbf{(Right)}: Comparison with imbalanced multimodal learning methods on the UCF101 dataset with 101 categories. Bold and underline mean the best and second-best results, respectively.}
\label{more}
\begin{tabular}{c|cc|cc}
\hline
Dataset       & \multicolumn{2}{c|}{CEFA}                            & \multicolumn{2}{c}{UCF101}                          \\ \hline
Methods       & \multicolumn{1}{c}{Acc}            & macro F1       & \multicolumn{1}{c}{Acc}            & macro F1       \\ \hline
Concatenation & \multicolumn{1}{c}{64.91}          & 65.26          & \multicolumn{1}{c}{80.41}          & 79.40          \\ 
Grad-Blending & \multicolumn{1}{c}{65.68}          & 66.11          & \multicolumn{1}{c}{81.73}          & 80.84          \\ 
OGM-GE        & \multicolumn{1}{c}{66.33}          & 66.76          & \multicolumn{1}{c}{81.15}          & 80.36          \\ 
AGM           & \multicolumn{1}{c}{67.65}          & 67.98          & \multicolumn{1}{c}{81.55}          & 80.36          \\ 
PMR           & \multicolumn{1}{c}{68.38}          & 68.89          & \multicolumn{1}{c}{81.36}          & 80.37          \\
MMPareto      & \multicolumn{1}{c}{71.18}          & 71.61          & \multicolumn{1}{c}{81.98}          & 80.64          \\ 
MLA           & \multicolumn{1}{c}{71.78}          & 72.02          & \multicolumn{1}{c}{82.01}          & \underline{81.22}          \\ 
D\&R          & \multicolumn{1}{c}{\underline{72.66}}          &\underline{ 72.97 }         & \multicolumn{1}{c}{\underline{82.11}}          & 80.87          \\ \hline
PDMP           & \multicolumn{1}{c}{\textbf{74.45}} & \textbf{74.78} & \multicolumn{1}{c}{\textbf{83.65}} & \textbf{82.08} \\ \hline
\end{tabular}

\end{table}

\textbf{Comparison in more-than-two modality case}. 
Most existing multimodal learning methods, such as G-Blending, OGM-GE, AGM, and PMR, primarily consider the scenarios involving only two modalities, which limits their generalizability to more complex real-world settings. In contrast, the proposed PDMP method is inherently flexible and scalable to accommodate an arbitrary number of modalities without additional structural adjustments. To demonstrate this advantage, we conduct comprehensive experiments on the CEFA dataset, which contains three distinct modalities—RGB, Depth, and IR.


To ensure fairness, we extend the core designs of baseline methods (e.g., OGM-GE, AGM, PMR, etc.) to operate in this tri-modal setting, just as D\&R do. As shown in the left part of Table~\ref{more}, PDMP significantly outperforms all competitors, achieving the highest accuracy (74.45\%) and macro F1 score (74.78\%) on CEFA. For instance, compared with the strongest baseline D\&R (72.66\% / 72.97\%), PDMP still delivers an absolute improvement of 1.79\% in accuracy and 1.81\% in macro F1, demonstrating its superior capability in effectively modeling and balancing multiple modalities.


\textbf{Comparison in large dataset}. Prior evaluations on datasets such as CRAME-D are limited in scale and complexity. To further assess the scalability and generalizability of PDMP, we conduct experiments on the challenging UCF101 dataset, which contains 101 categories.  As summarized in the right part of Table~\ref{more}, PDMP also demonstrates state-of-the-art performance, achieving the best results in both accuracy (83.65\%) and macro F1 score (82.08\%). Compared to strong baselines such as MLA (82.01\% / 81.22\%) and D\&R (82.11\% / 80.87\%), PDMP yields consistent performance gains, highlighting its robustness in handling complex multimodal data.

\subsection{Ablation study}
\label{ab}

\begin{table}[]
\centering
\renewcommand\arraystretch{1.1}
\caption{Performance on CREMA-D, AVE and KS datasets with various fusion methods. Combined with PDMP, conventional fusion methods consistently gain considerable improvement. † indicates PDMP is applied.}
\label{fusion_simple}
\setlength{\tabcolsep}{4mm}
\begin{tabular}{c|c|c|c}
\hline
Dataset       & CREMA-D       & AVE           & KS                  \\ \hline
Method        & Acc           & Acc           & Acc                   \\ \hline
Audio-only  &   57.27     &        62.16      &      48.67    \\
Visual-only   &   62.17      &       31.40        &    52.36       \\ \hline
Concatenation & 58.83          & 66.15          & 64.97           \\ 
Summation     & 62.24          & 67.25          & 64.53         \\ 
Film          & 56.21          & 59.18          & 57.35         \\ 
Gated         & 57.72          & 65.21          & 63.28         \\ \hline
Concatenation† & 80.21          & 71.28          & 72.18         \\ 
Summation†     & \textbf{82.54} & \textbf{71.65} & 73.17         \\ 
Film†          & 70.21         & 64.19          & 67.89          \\ 
Gated†         & 80.18          & 71.15          & \textbf{73.96} \\ \hline
\end{tabular}

\end{table}

\textbf{Comparison on conventional fusion methods}. In this experiment, we apply the PDMP strategy to four vanilla fusion methods: Concatenation, Summation, Film, and Gated. Among these, Summation is the type of late fusion method that fuses information at the logit level. The other three are the intermediate fusion methods that fuse information at the representation level

As shown in Table~\ref{fusion_simple}, the performance of each unimodal model in each dataset is inconsistent, as the audio performance is worse than the visual in CREMA-D, and on the contrary, the audio performance is better than the visual in AVE. This indicates that the performance-dominant modality could be different in the same modality combination task due to different scenes. Consequently, it is necessary to conduct the modality analysis for each dataset independently.

In addition, the accuracy of the visual-only model on the CREMA-D dataset is better than all of the vanilla fusion methods, which indicates that the learning of the multimodal model is indeed insufficient. After combining with PDMP, the performance of all the vanilla fusion methods consistently gains considerable improvement on all datasets, indicating the effectiveness and satisfactory flexibility of our method. In particular, for Summation fusion, PDMP achieves 20.30\%, 4.40\%, and  8.64\% accuracy improvement on the CRAME-D, AVE and KS datasets, respectively. This shows its superiority.




\textbf{Effect on different architectures}. To further verify the generalizability of the proposed PDMP strategy, we evaluate its performance on two widely used intermediate fusion methods: SE-Fusion\cite{surf} and mmFormer\cite{mmformer}, across all datasets. Unlike the previous four fusion methods that operate after the encoder or classifier, these two methods perform fusion within the modality encoder, offering a different fusion scenario.

As shown in Table~\ref{artecture}, applying the PDMP strategy to both SE-Fusion and mmFormer brings significant performance gains across all datasets. For example, SE-Fusion† improves the accuracy on CREMA-D from 61.22\% to 67.86\%, and on KS from 59.15\% to 65.27\%. Similarly, mmFormer† boosts the performance from 59.92\% to 68.37\% on CREMA-D, and from 64.26\% to 68.18\% on KS. This demonstrates that PDMP remains effective even when the fusion is conducted inside the encoder, highlighting its adaptability and potential for more complex multimodal fusion scenarios.

\begin{figure}[t]
\centering
\includegraphics[width=1.0\columnwidth]{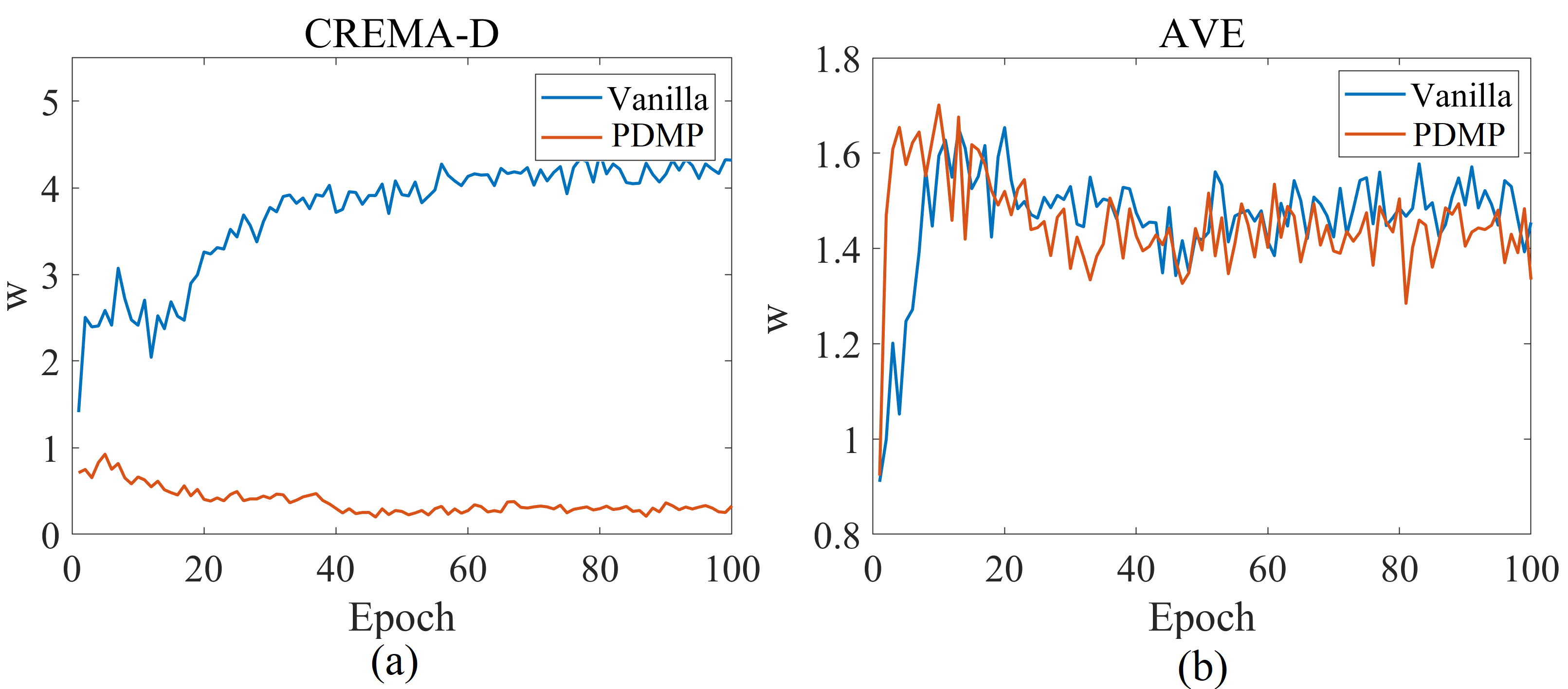}
\caption{ Visualization of optimization dependency coefficient $w$ on CREMA-D (a) and AVE (b) dataset, respectively. Here, $s_0$ and $s_1$ in Eq.~\ref{impact_new} represent the logit output of audio and visual modalities, respectively. `Vanilla' denotes the conventional multimodal model trained without PDMP.`PDMP' denotes the model trained with PDMP, respectively.}
\label{weight}
\end{figure}

\begin{table}[]
\centering
\renewcommand\arraystretch{1.1}
\caption{Performance on different datasets with CNN-based and transformer-based architectures, respectively. † indicates PDMP strategy is applied, which achieves better performance.}
\label{artecture}
\setlength{\tabcolsep}{4mm}
\begin{tabular}{c|c|c|c}
\hline
Dataset  & CREMA-D & AVE  & KS  \\ \hline
Method   & Acc     & Acc  & Acc  \\ \hline
SE-Fusion & 61.22    & 58.03 & 59.15 \\ 
SE-Fusion† & \textbf{67.86}    & \textbf{64.87} & \textbf{65.27} \\ \hline
mmFormer & 59.92    & 60.14 & 64.26  \\  
mmFormer† & \textbf{68.37}    & \textbf{67.49} & \textbf{68.18}\\ \hline
\end{tabular}

\end{table}

\textbf{Inter-modality performance comparison}. To further analyze PDMP, we visualize the change of the modality optimization dependency coefficient $w$ defined in Equation~\ref{impact_new}. As shown in Fig.~\ref{weight}, $w$ will rise quickly and then remain stable in both datasets when not using the PDMP strategy, which shows that the audio modality will dominate the optimization in the vanilla multimodal model. This is beneficial to AVE but not to the CRAME-D dataset since the audio modality is the performance-dominant mode for the AVE dataset but not the dominant modality for the CRAME-D dataset(see Table ~\ref{fusion_simple}). 


After applying the proposed PDMP strategy, the $w$ in  CRAME-D decays to a range less than 1, indicating that PDMP can help the performance-dominant modality dominate the optimization in multimodal learning. At the same time, $w$ with PDMP has a similar curve to that without PDMP in the AVE dataset, indicating that increasing the gradient for both modalities will not change the optimization-dominant modality in multimodal learning. These results show that the PDMP strategy can help the performance-dominant to dominate the optimization no matter whether it is the optimization-dominant modality or not.

\textbf{Analysis of different $\gamma_p$}. As described in Equation~\ref{gammar}, $\gamma_p$ determines the degree of gradient adjustment. Here, we discuss the effect of different $\gamma_p$ on the performance. As shown in Table~\ref{gamma_p}, the model achieves the best performance on CREMA-D, AVE, and CEFA datasets when $\gamma_p$ equals 15 and on KS when $\gamma_p$ equals 10. Besides, for all datasets, the performance first increases and then decreases after reaching the maximum value when $\gamma_p$ increases.  This is because when $\gamma_p$ increases, it accelerates the optimization of each modality encoder. However, if $\gamma_p$ is too large, the parameter update may be unstable due to high gradient and the model will also be under-optimized. 

\begin{table}[t]
\renewcommand\arraystretch{1.1}
\caption{Ablation experiments of  $\gamma_p$ on different datasets. The fusion method here is concatenation.}
\label{gamma_p}
\begin{tabular}{c|cccccc}
\hline
$\gamma_p$   & 1    & 5    & 10            & 15            & 20   & 25   \\ \hline
CREMA-D & 58.83 & 70.22 & 76.76          & \textbf{80.21} & 79.26 & 78.36 \\ 
AVE     & 66.15 & 67.55 & 69.73          & \textbf{71.28} & 70.64 & 69.85 \\ 
KS      & 64.97 & 68.39 & \textbf{72.18}          & 70.92 & 69.66 & 68.77 \\
\hline
\end{tabular}

\end{table}






\textbf{Analysis of the sampling number}. In the proposed PDMP method, the extra unimodal models to determine the modality property are the major additional cost. To reduce the computational burden, we consider utilizing subsets of large-scale datasets to analyze modality properties. Here, we use the KS dataset as an example, the results are shown in Table~\ref{sub}. When the scale of subset data is greater than 40\%, the relative performance difference of audio and visual modalities tends to be stable, indicating the subset of a large-scale dataset is enough to determine the modality property. Besides, the performance difference is not stable when the data scale is too small. This is because the visual modality contains more complex information and needs sufficient samples to optimize the model parameters.

\begin{table}[t]
\centering
\caption{Experiments on KS with different subset scales. n\% indicates the proportion of the training dataset. The fusion method here is concatenation.}
\label{sub}
\renewcommand\arraystretch{1.1}
\setlength{\tabcolsep}{3.0mm}{
\begin{tabular}{cccccc}
\hline
Scale  & 20\% & 40\% & 60\% & 80\% & 100\%         \\ \hline
Audio  & 26.9 & 31.8 & 36.4 & 44.6 & \textbf{48.7} \\ 
Visual & 22.7 & 36.7 & 43.2 & 49.7 & \textbf{52.6} \\ \hline
\end{tabular}}
\end{table}

\textbf{Effect of learning rate}. Considering that a larger $\gamma_p$ not only allows performance-dominant modality to dominate the optimization of the multimodal model, but it also speeds up the optimization of parameters. Therefore, we conduct experiments on different datasets to study the impact of the learning rate on the model optimization.

\begin{table}[t]
\centering
\renewcommand\arraystretch{1.1}
\caption{Ablation experiments of learning rate on different datasets.  † indicates PDMP is applied. $\gamma_r$ is set as 15}
\label{learning}
\setlength{\tabcolsep}{3mm}{
\begin{tabular}{c|cccc}
\hline
Dataset         & CREMA-D & AVE  & KS    \\ \hline
Learning Rate & Acc     & Acc  & Acc  \\ 
0.001         & 58.83    & 66.15 & 64.97 \\ 
0.015         & 63.45    & 69.16 & 68.13  \\ 
0.001†         & \textbf{80.21}    & \textbf{71.28} & \textbf{72.18}  \\ \hline
\end{tabular}
}

\end{table}

The results are shown in Table~\ref{learning}. We can see that although the learning rate has an impact on model performance, its contribution to the performance gain is limited. PDMP with $\gamma_r=15$ achieves better performance compared to directly increasing the learning rate by 15 times. Besides, the gradient modulation coefficient is not always greater than 1. For example, in the CRAME-D dataset, the modulation coefficient of the audio modality is 1/15, which is much smaller than 1. Therefore, the performance improvement brought by PMDP does not simply stem from an increase in the learning rate.

\textbf{Performance on large-scale datasets}. We conduct experiments on the VGGSound dataset to study the generalization ability of PDMP on the large-scale dataset.  Here, both the performance-dominant and optimization-dominant modalities are the audio modality, thus $\gamma_r=\gamma_p$. We set $\gamma_p$ as 15, which contributes to better performance. The unimodal and multimodal performance are listed in Table~\ref{vgg}. As can be seen from the table, PDMP achieves the best performance in both unimodal and multimodal settings. Specifically, its audio task performance (47.44\%) is 1.57\% higher than the second-ranked MLA, and its visual performance (32.65\%) is 1.05\% higher than MLA. Its multimodal performance (52.23\%) is also about 1.04\% and 1.39\% higher than MLA and D\&R, respectively.

\begin{table}[t]
    \centering
    \caption{Performance on the VGGSound dataset.}
    \label{vgg}
    \centering
    \setlength{\tabcolsep}{5.0mm}{
    \begin{tabular}{l|c|c|c}
        \hline
        Methods & Audio    & Visual   & Multi    \\
        \hline
        MLA     & 45.87    & 31.60    & 51.19    \\

        D\&R    & 45.18    & 31.23    & 50.84    \\

        PDMP    & \textbf{47.44} & \textbf{32.65} & \textbf{52.23} \\
        \hline
    \end{tabular}}
\end{table}

\section{Discussion}

\textbf{The computational cost of modality analysis}. Although the cost of modality analysis is inevitable, it is still within an acceptable range. First, as shown in Table~\ref{sub}, we can conduct modality analysis in the subsets of large-scale datasets to reduce the time cost. More importantly, the analysis results are reusable; only one modality analysis is required, and the performance evaluation of each modality can be applied to subsequent multimodal training. Compared with the cost of multimodal training itself, this part of the cost is almost negligible. More importantly, modality analysis may not be a necessary step in practical applications. Most tasks can avoid this part of the cost by reusing the results of existing unimodal experiments.

 \textbf{The result that  PDMP outperforms a naive learning rate increase in the ablation study on the AVE dataset}. The reason for this phenomenon may be attributed to the layer-wise learning rate dynamics. Due to gradient decay characteristics, parameters closer to the output layer typically receive higher gradients and consequently update faster than those closer to the input layer.    While increasing the learning rate could accelerate the update of input layers, it also affects the gradient of the output classifier and fusion module,  potentially leading to suboptimal convergence with excessive parameter oscillations. In contrast, PDMP specifically enhances gradients for the modal encoder near the input layer. This targeted approach accelerates training near the input layer while maintaining stable parameter updates at the output end, ultimately achieving superior performance.

\section{Conclusion}

In this paper, we re-analyze the under-optimization problem in multimodal learning and reveal that balanced learning is not the optimal strategy for multimodal learning. In contrast, the imbalanced learning dominated by the performance-dominant modality can contribute to better performance. Then we propose a simple but effective multimodal learning strategy called Performance-Dominant Modality Prioritization (PDMP) to alleviate the under-optimization problem, enabling the performance-dominant modality to dominate the optimization. This method achieves consistent performance gain on four representative multimodal datasets under various settings. In addition, PDMP can also generally serve as a flexible plug-in strategy for both CNN and Transformer-based models, demonstrating its practicality.


\bibliographystyle{IEEEtran}

\bibliography{IEEEabrv,IEEEexample}

@inproceedings{rgbd_seg1,
  title={Shapeconv: Shape-aware convolutional layer for indoor RGB-D semantic segmentation},
  author={Cao, Jinming and Leng, Hanchao and Lischinski, Dani and Cohen-Or, Daniel and Tu, Changhe and Li, Yangyan},
  booktitle={Proceedings of the IEEE/CVF International Conference on Computer Vision},
  pages={7088--7097},
  year={2021}
}

@inproceedings{rgbd_seg2,
  title={Acnet: Attention based network to exploit complementary features for rgbd semantic segmentation},
  author={Hu, Xinxin and Yang, Kailun and Fei, Lei and Wang, Kaiwei},
  booktitle={2019 IEEE International Conference on Image Processing (ICIP)},
  pages={1440--1444},
  year={2019},
  organization={IEEE}
}

@inproceedings{rgbd_seg3,
  title={Efficient rgb-d semantic segmentation for indoor scene analysis},
  author={Seichter, Daniel and K{\"o}hler, Mona and Lewandowski, Benjamin and Wengefeld, Tim and Gross, Horst-Michael},
  booktitle={2021 IEEE International Conference on Robotics and Automation (ICRA)},
  pages={13525--13531},
  year={2021},
  organization={IEEE}
}

@inproceedings{mm_detection2,
  title={Deep RGB-D saliency detection with depth-sensitive attention and automatic multi-modal fusion},
  author={Sun, Peng and Zhang, Wenhu and Wang, Huanyu and Li, Songyuan and Li, Xi},
  booktitle={Proceedings of the IEEE/CVF conference on computer vision and pattern recognition},
  pages={1407--1417},
  year={2021}
}

@article{mm_detection3,
  title={CDNet: Complementary depth network for RGB-D salient object detection},
  author={Jin, Wen-Da and Xu, Jun and Han, Qi and Zhang, Yi and Cheng, Ming-Ming},
  journal={IEEE Transactions on Image Processing},
  volume={30},
  pages={3376--3390},
  year={2021},
  publisher={IEEE}
}

@article{av1,
  title={Deep audio-visual speech recognition},
  author={Afouras, Triantafyllos and Chung, Joon Son and Senior, Andrew and Vinyals, Oriol and Zisserman, Andrew},
  journal={IEEE transactions on pattern analysis and machine intelligence},
  volume={44},
  number={12},
  pages={8717--8727},
  year={2018},
  publisher={IEEE}
}

@article{av2,
  title={Multimodal sparse transformer network for audio-visual speech recognition},
  author={Song, Qiya and Sun, Bin and Li, Shutao},
  journal={IEEE Transactions on Neural Networks and Learning Systems},
  year={2022},
  publisher={IEEE}
}

@inproceedings{ar1,
  title={Learnable irrelevant modality dropout for multimodal action recognition on modality-specific annotated videos},
  author={Alfasly, Saghir and Lu, Jian and Xu, Chen and Zou, Yuru},
  booktitle={Proceedings of the IEEE/CVF Conference on Computer Vision and Pattern Recognition},
  pages={20208--20217},
  year={2022}
}

@inproceedings{ogm,
  title={Balanced multimodal learning via on-the-fly gradient modulation},
  author={Peng, Xiaokang and Wei, Yake and Deng, Andong and Wang, Dong and Hu, Di},
  booktitle={Proceedings of the IEEE/CVF Conference on Computer Vision and Pattern Recognition},
  pages={8238--8247},
  year={2022}
}

@article{umt,
  title={Improving multi-modal learning with uni-modal teachers},
  author={Du, Chenzhuang and Li, Tingle and Liu, Yichen and Wen, Zixin and Hua, Tianyu and Wang, Yue and Zhao, Hang},
  journal={arXiv preprint arXiv:2106.11059},
  year={2021}
}

@inproceedings{wh,
  title={What makes training multi-modal classification networks hard?},
  author={Wang, Weiyao and Tran, Du and Feiszli, Matt},
  booktitle={Proceedings of the IEEE/CVF Conference on Computer Vision and Pattern Recognition},
  pages={12695--12705},
  year={2020}
}

@article{Crema,
  title={Crema-d: Crowd-sourced emotional multimodal actors dataset},
  author={Cao, Houwei and Cooper, David G and Keutmann, Michael K and Gur, Ruben C and Nenkova, Ani and Verma, Ragini},
  journal={IEEE transactions on affective computing},
  volume={5},
  number={4},
  pages={377--390},
  year={2014},
  publisher={IEEE}
}

@article{soomro2012ucf101,
  title={UCF101: A dataset of 101 human actions classes from videos in the wild},
  author={Soomro, K. and Zamir, A. R. and Shah, M.},
  year={2012},
  journal={arXiv preprint arXiv:1212.0402}
}

@inproceedings{pmr,
  title={PMR: Prototypical Modal Rebalance for Multimodal Learning},
  author={Fan, Yunfeng and Xu, Wenchao and Wang, Haozhao and Wang, Junxiao and Guo, Song},
  booktitle={Proceedings of the IEEE/CVF Conference on Computer Vision and Pattern Recognition},
  pages={20029--20038},
  year={2023}
}

@inproceedings{MARS,
  title={Mars: Motion-augmented rgb stream for action recognition},
  author={Crasto, Nieves and Weinzaepfel, Philippe and Alahari, Karteek and Schmid, Cordelia},
  booktitle={Proceedings of the IEEE/CVF Conference on Computer Vision and Pattern Recognition},
  pages={7882--7891},
  year={2019}
}

@inproceedings{MH2,
  title={Modality distillation with multiple stream networks for action recognition},
  author={Garcia, Nuno C and Morerio, Pietro and Murino, Vittorio},
  booktitle={Proceedings of the European Conference on Computer Vision},
  pages={103--118},
  year={2018}
}

@article{MH3,
  title={Dynamic-Hierarchical Attention Distillation With Synergetic Instance Selection for Land Cover Classification Using Missing Heterogeneity Images},
  author={Li, Xiao and Lei, Lin and Sun, Yuli and Kuang, Gangyao},
  journal={IEEE Transactions on Geoscience and Remote Sensing},
  volume={60},
  pages={1--16},
  year={2021},
  publisher={IEEE}
}

@inproceedings{AVE,
  title={Audio-visual event localization in unconstrained videos},
  author={Tian, Yapeng and Shi, Jing and Li, Bochen and Duan, Zhiyao and Xu, Chenliang},
  booktitle={Proceedings of the European conference on computer vision (ECCV)},
  pages={247--263},
  year={2018}
}

@inproceedings{librosa,
  title={librosa: Audio and music signal analysis in python},
  author={McFee, Brian and Raffel, Colin and Liang, Dawen and Ellis, Daniel P and McVicar, Matt and Battenberg, Eric and Nieto, Oriol},
  booktitle={Proceedings of the 14th python in science conference},
  volume={8},
  pages={18--25},
  year={2015}
}

@article{mmformer,
  title={mmFormer: Multimodal Medical Transformer for Incomplete Multimodal Learning of Brain Tumor Segmentation},
  author={Zhang, Yao and He, Nanjun and Yang, Jiawei and Li, Yuexiang and Wei, Dong and Huang, Yawen and Zhang, Yang and He, Zhiqiang and Zheng, Yefeng},
  journal={arXiv preprint arXiv:2206.02425},
  year={2022}
}

@inproceedings{surf,
  title={A dataset and benchmark for large-scale multi-modal face anti-spoofing},
  author={Zhang, Shifeng and Wang, Xiaobo and Liu, Ajian and Zhao, Chenxu and Wan, Jun and Escalera, Sergio and Shi, Hailin and Wang, Zezheng and Li, Stan Z},
  booktitle={Proceedings of the IEEE/CVF Conference on Computer Vision and Pattern Recognition},
  pages={919--928},
  year={2019}
}

@inproceedings{cefa,
  title={Casia-surf cefa: A benchmark for multi-modal cross-ethnicity face anti-spoofing},
  author={Liu, Ajian and Tan, Zichang and Wan, Jun and Escalera, Sergio and Guo, Guodong and Li, Stan Z},
  booktitle={Proceedings of the IEEE/CVF Winter Conference on Applications of Computer Vision},
  pages={1179--1187},
  year={2021}
}

@inproceedings{ks,
  title={Look, listen and learn},
  author={Arandjelovic, Relja and Zisserman, Andrew},
  booktitle={Proceedings of the IEEE international conference on computer vision},
  pages={609--617},
  year={2017}
}

@inproceedings{mem1,
  title={A closer look at memorization in deep networks},
  author={Arpit, Devansh and Jastrz{{e}}bski, Stanis{\l}aw and Ballas, Nicolas and Krueger, David and Bengio, Emmanuel and Kanwal, Maxinder S and Maharaj, Tegan and Fischer, Asja and Courville, Aaron and Bengio, Yoshua and others},
  booktitle={International conference on machine learning},
  pages={233--242},
  year={2017},
  organization={PMLR}
}

@article{mem2,
  title={Co-teaching: Robust training of deep neural networks with extremely noisy labels},
  author={Han, Bo and Yao, Quanming and Yu, Xingrui and Niu, Gang and Xu, Miao and Hu, Weihua and Tsang, Ivor and Sugiyama, Masashi},
  journal={Advances in neural information processing systems},
  volume={31},
  year={2018}
}

@article{mem3,
  title={Understanding deep learning (still) requires rethinking generalization},
  author={Zhang, Chiyuan and Bengio, Samy and Hardt, Moritz and Recht, Benjamin and Vinyals, Oriol},
  journal={Communications of the ACM},
  volume={64},
  number={3},
  pages={107--115},
  year={2021},
  publisher={ACM New York, NY, USA}
}

@inproceedings{mmcosine,
  title={MMCosine: Multi-Modal Cosine Loss Towards Balanced Audio-Visual Fine-Grained Learning},
  author={Xu, Ruize and Feng, Ruoxuan and Zhang, Shi-Xiong and Hu, Di},
  booktitle={ICASSP 2023-2023 IEEE International Conference on Acoustics, Speech and Signal Processing (ICASSP)},
  pages={1--5},
  year={2023},
  organization={IEEE}
}

@article{FMV,
  title={Enhancing Multi-modal Cooperation via Fine-grained Modality Valuation},
  author={Wei, Yake and Feng, Ruoxuan and Wang, Zihe and Hu, Di},
  journal={arXiv preprint arXiv:2309.06255},
  year={2023}
}

@article{MLA,
  title={Multimodal representation learning by alternating unimodal adaptation},
  author={Zhang, Xiaohui and Yoon, Jaehong and Bansal, Mohit and Yao, Huaxiu},
  journal={arXiv preprint arXiv:2311.10707},
  year={2023}
}

@inproceedings{agm,
  title={Boosting Multi-modal Model Performance with Adaptive Gradient Modulation},
  author={Li, Hong and Li, Xingyu and Hu, Pengbo and Lei, Yinuo and Li, Chunxiao and Zhou, Yi},
  booktitle={Proceedings of the IEEE/CVF International Conference on Computer Vision},
  pages={22214--22224},
  year={2023}
}

@inproceedings{diagnosing,
  title={Diagnosing and Re-learning for Balanced Multimodal Learning},
  author={Wei, Yake and Li, Siwei and Feng, Ruoxuan and Hu, Di},
  booktitle={European Conference on Computer Vision},
  pages={71--86},
  year={2025},
  organization={Springer}
}

@inproceedings{vggsound,
  title={Vggsound: A large-scale audio-visual dataset},
  author={Chen, Honglie and Xie, Weidi and Vedaldi, Andrea and Zisserman, Andrew},
  booktitle={ICASSP 2020-2020 IEEE International Conference on Acoustics, Speech and Signal Processing (ICASSP)},
  pages={721--725},
  year={2020},
  organization={IEEE}
}

@article{tmm-c1,
  title={Unimf: A unified multimodal framework for multimodal sentiment analysis in missing modalities and unaligned multimodal sequences},
  author={Huan, Ruohong and Zhong, Guowei and Chen, Peng and Liang, Ronghua},
  journal={IEEE Transactions on Multimedia},
  volume={26},
  pages={5753--5768},
  year={2023},
  publisher={IEEE}
}

@article{tmm-c3, title={Dynamically shifting multimodal representations via hybrid-modal attention for multimodal sentiment analysis}, author={Lin, Ronghao and Hu, Haifeng}, journal={IEEE Transactions on Multimedia}, volume={26}, pages={2740--2755}, year={2023}, publisher={IEEE} }

@article{tmm-c2, title={Image-text multimodal emotion classification via multi-view attentional network}, author={Yang, Xiaocui and Feng, Shi and Wang, Daling and Zhang, Yifei}, journal={IEEE Transactions on Multimedia}, volume={23}, pages={4014--4026}, year={2020}, publisher={IEEE} }

@article{reconboost,
  title={ReconBoost: Boosting Can Achieve Modality Reconcilement},
  author={Hua, Cong and Xu, Qianqian and Bao, Shilong and Yang, Zhiyong and Huang, Qingming},
  journal={arXiv preprint arXiv:2405.09321},
  year={2024}
}

@article{mmpareto,
  title={MMPareto: Boosting Multimodal Learning with Innocent Unimodal Assistance},
  author={Wei, Yake and Hu, Di},
  journal={arXiv preprint arXiv:2405.17730},
  year={2024}
}

@book{cover_thomas,
  title={Elements of Information Theory},
  author={Cover, Thomas M. and Thomas, Joy A.},
  year={2006},
  publisher={Wiley}
}

@article{tishby1999ib,
  title={The Information Bottleneck Method},
  author={Tishby, Naftali and Pereira, Fernando C. and Bialek, William},
  journal={arXiv preprint physics/0004057},
  year={1999}
}

@inproceedings{zhang2025gmml, title={GMML: Gradient-Modulated Robustness for Imbalance-Aware Multimodal Learning}, author={Zhang, Zikai and Zhang, Xu and Li, Ziyi and Li, Yidong and Cao, Yuanzhouhan}, booktitle={Proceedings of the 33rd ACM International Conference on Multimedia}, pages={7922--7930}, year={2025} }

@inproceedings{hu2025geometric,
  title={Geometric Gradient Divergence Modulation for Imbalanced Multimodal Learning},
  author={Hu, Disen and Jiang, Xun and Sun, Zhe and Yang, Hao and Peng, Chong and Yan, Peng and Shen, Heng Tao and Xu, Xing},
  booktitle={Proceedings of the 33rd ACM International Conference on Multimedia},
  pages={1337--1345},
  year={2025}
}

@article{guo2024classifier,
  title={Classifier-guided gradient modulation for enhanced multimodal learning},
  author={Guo, Zirun and Jin, Tao and Chen, Jingyuan and Zhao, Zhou},
  journal={Advances in Neural Information Processing Systems},
  volume={37},
  pages={133328--133344},
  year={2024}
}

@inproceedings{dgl,
  title={Boosting Multimodal Learning via Disentangled Gradient Learning},
  author={Wei, Shicai and Luo, Chunbo and Luo, Yang},
  booktitle={Proceedings of the IEEE/CVF International Conference on Computer Vision},
  pages={22879--22888},
  year={2025}
}

@inproceedings{arl, title={Improving multimodal learning via imbalanced learning}, author={Wei, Shicai and Luo, Chunbo and Luo, Yang}, booktitle={Proceedings of the IEEE/CVF International Conference on Computer Vision}, pages={2250--2259}, year={2025} }

@inproceedings{LFM,
  title={Facilitating multimodal classification via dynamically learning modality gap},
  author={Yang, Y. and Wan, F. and Jiang, Q. Y. and Xu, Y.},
  booktitle={Advances in Neural Information Processing Systems},
  volume={37},
  pages={62108--62122},
  year={2024}
}

@inproceedings{Wei2023MMANet,
  author    = {Wei, Shicai and Luo, Chunbo and Luo, Yang},
  title     = {{MMANet}: Margin-Aware Distillation and Modality-Aware Regularization for Incomplete Multimodal Learning},
  booktitle = {Proceedings of the IEEE/CVF Conference on Computer Vision and Pattern Recognition (CVPR)},
  pages     = {20039--20049},
  year      = {2023},
  doi       = {10.1109/CVPR52729.2023.01919},
  url       = {https://doi.org/10.1109/CVPR52729.2023.01919}
}

@article{Wei2023MSHNet,
  author  = {Wei, Shicai and Luo, Yang and Ma, Xiaoguang and Ren, Peng and Luo, Chunbo},
  title   = {{MSH-Net}: Modality-Shared Hallucination With Joint Adaptation Distillation for Remote Sensing Image Classification Using Missing Modalities},
  journal = {IEEE Transactions on Geoscience and Remote Sensing},
  volume  = {61},
  pages   = {1--15},
  year    = {2023},
  doi     = {10.1109/TGRS.2023.3265650},
  url     = {https://doi.org/10.1109/TGRS.2023.3265650}
}

@inproceedings{Wei2024ScaledDecoupledDistillation,
  author    = {Wei, Shicai and Luo, Chunbo and Luo, Yang},
  title     = {Scaled Decoupled Distillation},
  booktitle = {Proceedings of the IEEE/CVF Conference on Computer Vision and Pattern Recognition (CVPR)},
  pages     = {15975--15983},
  year      = {2024},
  doi       = {10.1109/CVPR52733.2024.01512},
  url       = {https://doi.org/10.1109/CVPR52733.2024.01512}
}

@article{Wei2024PrivilegedModalityLearning,
  author  = {Wei, Shicai and Luo, Chunbo and Luo, Yang and Xu, Jialang},
  title   = {Privileged Modality Learning via Multimodal Hallucination},
  journal = {IEEE Transactions on Multimedia},
  volume  = {26},
  pages   = {1516--1527},
  year    = {2024},
  doi     = {10.1109/TMM.2023.3282874},
  url     = {https://doi.org/10.1109/TMM.2023.3282874}
}

@article{Wei2024GradientDecoupledLearning,
  author  = {Wei, Shicai and Luo, Chunbo and Ma, Xiaoguang and Luo, Yang},
  title   = {Gradient Decoupled Learning With Unimodal Regularization for Multimodal Remote Sensing Classification},
  journal = {IEEE Transactions on Geoscience and Remote Sensing},
  volume  = {62},
  pages   = {1--12},
  year    = {2024},
  doi     = {10.1109/TGRS.2024.3478393},
  url     = {https://doi.org/10.1109/TGRS.2024.3478393}
}

@article{Wei2023DiversityGuidedDistillation,
  author  = {Wei, Shicai and Luo, Yang and Luo, Chunbo},
  title   = {Diversity-Guided Distillation With Modality-Center Regularization for Robust Multimodal Remote Sensing Image Classification},
  journal = {IEEE Transactions on Geoscience and Remote Sensing},
  volume  = {61},
  pages   = {1--14},
  year    = {2023},
  doi     = {10.1109/TGRS.2023.3336297},
  url     = {https://doi.org/10.1109/TGRS.2023.3336297}
}

@article{Wei2023OneStageModalityDistillation,
  author  = {Wei, Shicai and Luo, Yang and Luo, Chunbo},
  title   = {One-stage Modality Distillation for Incomplete Multimodal Learning},
  journal = {CoRR},
  volume  = {abs/2309.08204},
  year    = {2023},
  doi     = {10.48550/arXiv.2309.08204},
  url     = {https://doi.org/10.48550/arXiv.2309.08204}
}

\end{document}